\documentclass{article}

\usepackage{microtype}
\usepackage{graphicx}
\usepackage{subfigure}
\usepackage{booktabs} 

\usepackage{hyperref}

\usepackage{balance}
\usepackage{natbib}
\usepackage[english]{babel}
\usepackage{latexsym}
\usepackage{amsmath}

\usepackage{amssymb}
\usepackage{amsthm}
\usepackage{xspace}
\usepackage{algpseudocode}

\usepackage[disable]{todonotes} %

\newlength\myindent
\algdef{SE}[SUBALG]{Indent}{EndIndent}{}{\algorithmicend\ }%
\algtext*{Indent}
\algtext*{EndIndent}

\usepackage[capitalize]{cleveref}

\newtheorem{theorem}{Theorem}

\newtheorem{lemma}[theorem]{Lemma}
\newtheorem{corollary}[theorem]{Corollary}
\newtheorem{definition}[theorem]{Definition}

\usepackage[accepted]{icml2018}

\algnewcommand{\Initialize}[1]{%
  \State \textbf{Initialize:}
  \Statex \hspace*{\algorithmicindent}\parbox[t]{.8\linewidth}{\raggedright #1}
}
\algnewcommand{\Inputs}[1]{%
  \State \textbf{Inputs:}
  \Statex \hspace*{\algorithmicindent}\parbox[t]{.8\linewidth}{\raggedright #1}
}

\newcommand{\RuntimeEst}{{\sc RuntimeEst}\xspace}
\newcommand{\LeapsAndBounds}{{\sc LeapsAndBounds}\xspace}
\DeclareMathOperator{\argmin}{argmin}

\newcommand{\lb}{\mbox{LB}}
\newcommand{\ordo}{\mathcal{O}}
\newcommand{\opt}{\mathrm{ OPT}\xspace}

\newcommand{\minisat}{\texttt{minisat}\xspace}
 
\newcommand{\E}{\mathbb E}
\newcommand{\N}{\mathcal{N}}
\newcommand{\ed}{$(\epsilon, \delta)$\xspace}
\newcommand{\defeq}{\doteq}
\newcommand{\Jlist}{\mathcal{J}}

\renewcommand{\epsilon}{\varepsilon}
\newcommand{\eps}{\varepsilon}
\newcommand{\ceil}[1]{\left\lceil {#1} \right\rceil}
\newcommand{\floor}[1]{\left\lfloor {#1} \right\rfloor}

\makeatletter
\newcommand\footnoteref[1]{\protected@xdef\@thefnmark{\ref{#1}}\@footnotemark}
\makeatother

\begin{document}

\icmltitlerunning{\LeapsAndBounds: A Method for Approximately Optimal Algorithm Configuration}

\twocolumn[
\icmltitle{\LeapsAndBounds: A Method for Approximately Optimal Algorithm Configuration}




\begin{icmlauthorlist}
\icmlauthor{Gell\'ert Weisz}{dm}
\icmlauthor{Andr\'as Gy\"orgy}{dm,ic}
\icmlauthor{Csaba Szepesv\'ari}{dm,ua}
\end{icmlauthorlist}

\icmlaffiliation{dm}{DeepMind, London, UK.}
\icmlaffiliation{ic}{On leave from Imperial College London, London, UK.}
\icmlaffiliation{ua}{On leave from University of Alberta, Edmonton, AB, Canada}

\icmlcorrespondingauthor{Gell\'ert Weisz}{gellert@google.com}
\icmlcorrespondingauthor{Andr\'as Gy\"orgy}{agyorgy@google.com}
\icmlcorrespondingauthor{Csaba Szepesv\'ari}{szepi@google.com}

\icmlkeywords{algorithm configuration, sampling, Bernstein stopping, SAT solver}

\vskip 0.3in
]



\printAffiliationsAndNotice{}  

\begin{abstract}
We consider the problem of configuring general-purpose solvers to run efficiently on 
problem instances drawn from an unknown distribution.
The goal of the configurator is to find a configuration that runs fast on average on most instances, and do so with the least amount of total work. It can run a chosen solver on a random instance until the solver finishes
or a timeout is reached. 
We propose \LeapsAndBounds, an algorithm that tests configurations on
randomly selected problem instances for longer and longer time.
We prove that the capped expected runtime of the configuration returned by \LeapsAndBounds
is close to the optimal expected runtime, while our algorithm's running time is near-optimal.
Our results show that \LeapsAndBounds is more efficient than the recent algorithm of \citet{kleinberg2017efficiency}, which, 
to our knowledge, is the only other algorithm configuration method with non-trivial theoretical guarantees. 
Experimental results on configuring a public SAT solver on a new benchmark dataset also stand witness to 
the superiority of our method.
\end{abstract}

\section{Introduction}

For computational problems of major practical interest (satisfiability, planning, etc.) the computing science community
has developed a large number of highly configurable ``solvers.''
The reason is that while the hardest problem instances take a long time to solve by any of the solvers, 
the instances that one encounters in practical applications may exhibit specific properties so that
the appropriate solver with an appropriate configuration may finish much faster.
The plethora of solvers and their configurations, which for simplicity of presentation we will just treat as configurations from this point on, is explained by the diversity of applications.
Which configuration to use in a specific application can then be treated as a learning problem, 
where an application is identified with an unknown distribution over problem instances that one can sample from, the learning algorithm can run any configuration on any sampled instance until a timeout of its choice,
 and the goal is to find a configuration with nearly optimal expected runtime while using the least amount of time during the search.%
\footnote{Related, but different problems are considered, e.g., by \citet{Luby93optimalspeedup,Adam01learningwhile,mnih2008empirical,AB10,metamax,Li16hyperband}.} 
There has been much practical success on designing such black-box configuration search methods, especially in the context of satisfiability problems.
Examples of successful methods include
ParamILS \cite{hutter2007potential,hutter2009paramils}, SMAC \cite{hutter2011sequential,hutter2013bayesian}, irace \cite{birattari2002racing,lopez2011irace}, and GGA \cite{ansotegui2009gender,ansotegui2015model}.
These methods themselves rely on many heuristics and as such
lack theoretical guarantees.

Recently, \citet{kleinberg2017efficiency} explored this problem, presenting a general-purpose configuration optimizer called \emph{Structured Procrastination}, 
with guarantees on both {\em (i)} how close to the optimal configuration the algorithm's result is, and {\em (ii)} how long it takes to find such a configuration. For {\em (ii)}, \citet{kleinberg2017efficiency} prove that the expected runtime of their algorithm is within a logarithmic factor of the optimal runtime in a worst-case sense. 
Furthermore, they show that the gap between worst-case runtimes of existing algorithms (SMAC, ROAR, ParamILS, GGA, irace) and their solution can be arbitrarily large. Structured Procrastination attempts to refine the runtime guarantee for the empirically fastest solver and solves tasks in increasing order of difficulty, 
postponing difficult tasks until all simpler tasks have been solved.
The main novelty of their work is that it comes with theoretical guarantees (lower and upper bounds on the runtime),
but no empirical illustration is provided. 

This paper builds on the results of \citet{kleinberg2017efficiency}, and our problem statement closely follows theirs. Our main technical contributions 
are as follows: We present an (arguably simpler) algorithm (\LeapsAndBounds) that finds an approximately optimal configuration with a worst-case runtime bound that improves upon that of \citet{kleinberg2017efficiency}, while we consider a broader class of problems (we don not need their global runtime cap). 
We also present instance-dependent runtime bounds that show that
\LeapsAndBounds finishes faster if the runtime of the configurations over different problem instances has small variance. Experiments were carried out to assess practical performance of both Structured Procrastination and \LeapsAndBounds on configuring the open-source \minisat solver. \LeapsAndBounds runs every configuration for less time than Structured Procrastination, and returns significantly faster. 
Finally, to facilitate further research and enable direct comparison to our results, our large-scale measurements on running times of the \minisat solver 
are published together with the paper.%
\footnote{\url{https://github.com/deepmind/leaps-and-bounds}}

The rest of the paper is organized as follows: The problem is introduced formally in \cref{sec:problem}. For clarity, the most basic version of our algorithm is presented first in \cref{sec:alg}, and its performance is analyzed in \cref{sec:analysis}. Improvements to our method, together with their analyses, are presented in \cref{sec:empirical}.
Experimental results are presented in \cref{sec:exp}, followed by some notes on parallel implementation in \cref{sec:parallel}. Finally, conclusions are drawn in \cref{sec:conc}.

\vspace{-6pt}

\section{Problem Statement}
\label{sec:problem}

Following \citet{kleinberg2017efficiency}, 
the algorithm configuration problem is defined by a tuple $(\N, \Gamma, R, \kappa_0)$ as follows:%
\footnote{Compared to their problem statement, we removed the global runtime cap from the definition as it is not required for our results.}
Here, $\N$ is a family of configurations and 
$\Gamma$ is a distribution over input instances.\footnote{For randomized solvers, input instances can mean (input instance, random seed) pairs.} For now, we consider the case when $\N$ is a finite set.
If we have a benchmark set of instances, 
	we let $\Gamma$ be the uniform distribution over these benchmark instances. 
For configuration $i\in \N$ and instance $j$,
	$R(i, j)\in [0,\infty]$ is the execution time of configuration $i$ on instance $j$. 
Finally, $\kappa_0>0$ is the minimum runtime: For all $i,j$ pairs, $R(i, j)\ge \kappa_0$. 

We let $R(i) = \E_{J\sim\Gamma}\left[ R(i, J) \right]$ denote the average runtime of configuration $i$ on instances distributed according to $\Gamma$, and define $\opt=\min_i R(i)$ as the mean runtime of an optimal configuration.
\emph{Our goal is to find such an optimal, or at least nearly optimal configuration 
while spending as little time as possible--proportional to the runtime of the optimal configuration--on this task.}
For this, a search algorithm can {\em (i)} sample instances $J$ at random from $\Gamma$; {\em (ii)} enumerate the configurations in $\N$; {\em (iii)} run a configuration $i$ on an instance $j$ until it finishes, or the execution time exceeds a fixed timeout $\tau\ge 0$, chosen by the search algorithm. Practically, this means observing 
$R(i,j,\tau) \defeq \min( R(i,j), \tau )$ after time $R(i,j,\tau)$, and also whether the calculation has finished with a solution or it timed out. 

The main difficulty in organizing the search is that some configurations may take a long, or even infinite time to execute
on some instances.
Since an algorithm that claims to \emph{find} a near-optimal configuration must verify 
that no other configuration can finish significantly faster than the chosen configuration, the total runtime
is at least proportional to $n \times \opt$, where $n=|N|$ is the number of configurations to be tested.
Since knowing the mean runtime up to a multiplicative accuracy of $(1+\epsilon)$ requires $\Omega(1/\epsilon^2)$ samples
even when the runtime distributions are light-tailed,
relaxing the requirement to find a configuration $i$ with runtime $R(i)\le (1+\epsilon)\opt$,
we get that the total runtime is at least $\Omega(n\times \opt/\epsilon^2)$.
The situation worsens for heavy-tailed runtime distributions:
If the runtime of an algorithm is $b>1$ with probability $1/b$ and
 $0$ otherwise, with $b$ unknown, all sampling methods need to see at least
 one positive runtime to estimate the expected runtime up to any fixed accuracy. Thus, any sampling method
 needs to use at least $\Omega(b)$ time, despite that the expected runtime is constant $1$ (independently of $b$). 
This implies that in the face of heavy-tailed runtime distributions,
the runtime of any sound configuration search algorithm would be unbounded in the worst-case,
regardless the value of $\opt$, $n$ and $\eps$.
Since heavy tailed runtime distributions are quite common in practice, rather than constraining the problem
by ruling these out,
following \citet{kleinberg2017efficiency},
we relax the search criterion to that of finding an $(\epsilon,\delta)$-optimal configuration.
Introducing the $\tau$-capped version of $R(i)$ as 
$R_\tau(i) = \E_{J\sim\Gamma}\left[ R(i, J, \tau) \right]$, we have the following definition for 
 $(\epsilon,\delta)$-optimality:
\begin{definition}[\ed-optimality]
\label{def:edopt}
A configuration $i^*$ is \ed-optimal if there exists some threshold $\tau$ such that $R_\tau(i^*) \leq (1+\epsilon)\opt$, and $\Pr_{J\sim\Gamma} \left(R(i^*, J) > \tau\right) \leq \delta$. Otherwise, we say $i^*$ is \ed-suboptimal.
\end{definition}
In words, given \ed, a sound configuration search algorithm must find a configuration $i$ whose $\tau$-capped mean runtime is at most $(1+\eps)\opt $, with a $\tau$ 
larger than 
the $\delta$-quantile of the runtime distribution of configuration $i$. 
This is a reasonable criterion when $\opt$ is reasonably small.\footnote{If some problem instances are hopelessly hard, the expected runtime of even an optimal configuration can be infinite, in which case any configuration becomes \ed-optimal. To alleviate this problem, it would be more meaningful to define \ed-optimality with respect to the optimal capped runtime; this is left for future work (see \cref{sec:conc} for more details).}

In this paper we introduce the algorithm \LeapsAndBounds that identifies an \ed-optimal configuration with probability $1-\zeta$ for a failure parameter $\zeta$ and has an expected runtime of $\ordo\left(\opt \frac{n}{\epsilon^2\delta} \log\big(\frac{n\log \opt}{\zeta}\big)\right)$. The method of \citet{kleinberg2017efficiency} has an additional assumption that all runtimes of any configuration on any instance sampled from $\Gamma$ are below a maximum $\bar{\kappa}$ that must also be known by the algorithm. 
While this renders the runtime distributions light-tailed, a nice feature of their method 
is that its runtime
$\ordo\!\left(\opt \frac{n}{\epsilon^2\delta} \log\big(\frac{n\log \bar{\kappa}}{\zeta\delta\epsilon^2}\big)\right)$ has only a mild dependence on $\bar\kappa$.
\LeapsAndBounds does not require a runtime cap and we shave off a few terms from their bound:
We replace the doubly  logarithmic dependence on $\bar{\kappa}$ with an identical dependence 
on the practically much smaller $\opt$, and remove logarithmic terms that depend on
 $\delta^{-1}$ and $\epsilon^{-2}$.
\citet{kleinberg2017efficiency} also prove that the minimum worst-case runtime for any algorithm is $\Omega\left(\opt \frac{n}{\epsilon^2\delta} \right)$, so both methods are within a logarithmic factor of the optimum.

The above results make sense when $n=|N|$ is small enough to allow running each algorithm configuration. Similarly to \citet{kleinberg2017efficiency}, \LeapsAndBounds can be extended to the case of an arbitrarily large number of configurations: by sampling $n$ configurations randomly from the set of all configurations, the probability that none of the fastest $\gamma$ fraction of configurations have been sampled is at most $C e^{-n\gamma}$ for a universal constant $C>0$. Thus, by letting $n=\ceil{\frac{1}{\gamma}\log(C/\zeta)}$, with probability $1-2\zeta$, \LeapsAndBounds returns an \ed-optimal configuration with respect to a configuration from the fastest $\gamma$ fraction of configurations from the entire space.

\section{Algorithm}
\label{sec:alg}

The main problem in finding a near-optimal solver configuration is that solving some instances may take arbitrarily long. To alleviate the problem, \ed-optimality only considers the mean of runtimes capped at a timeout, ensuring that at most a $\delta$ fraction of the worst instances run longer than this timeout. This makes estimating the average runtime of a configuration (over random instances) possible through sampling. The main issue with sampling is that computing the average runtime over the samples can be slowed down arbitrarily if we accidentally select a problem instance with a very large running time. This could be avoided if an oracle told us the runtime threshold $\tau$ in the definition of \ed-optimality, but this is not available of course.
To solve the problem, we present a configuration optimization algorithm called \LeapsAndBounds (\cref{master-alg}).

\setlength{\textfloatsep}{18pt}
\begin{algorithm}[t]
\caption{\LeapsAndBounds}\label{master-alg}
\begin{algorithmic}[1]
\Inputs{
  Set $\N$ of $n$ algorithm configurations\\
  Precision parameter $\epsilon \in (0, \frac{1}{3})$\\
  Quantile parameter $\delta\in(0, 1)$\\
  Failure probability parameter $\zeta\in(0, 1)$\\
  Lower runtime bound $\kappa_0>0$\\
  Instance distribution $\Gamma$\\
}
\Initialize{$\theta \gets \frac{16}{7}\kappa_0$, $k\gets 0$, $\Jlist\gets$ empty list}
\While{True}
  \State $k\gets k+1$ \Comment{phase count}
  \State $b\gets \ceil{44\log\left(\frac{6nk(k+1)}{\zeta}\right)\frac{1}{\delta\epsilon^2}}$
  	\Comment{instance bound}
  \State Add $b-|\Jlist|$ new instances sampled from $\Gamma$ to $\Jlist$ \label{alg:master:sample}
  \For{$i \in \N$} \label{alg:master:for}
    \State $\bar{Q}_i \gets$\RuntimeEst$(i, \Jlist, \delta, \theta)$ \label{alg:master:runtimeest}
  \EndFor
  \If{$\min_i \bar{Q}_i < \theta$}
    \State \Return $\argmin_i \bar{Q}_i$
  \EndIf
  \State $\theta \gets 2\theta$ \label{alg:master:doubling}
\EndWhile
\end{algorithmic}
\end{algorithm}

\LeapsAndBounds attempts to guess a rough value of OPT, starting from a low value. Calling its guess $\theta$, the algorithm then tries to find a configuration with a mean runtime less than $\theta$. If this succeeds, it returns the configuration with the smallest mean found. Otherwise, $\theta$ is doubled and a new phase is started.
The simplest way of measuring the mean runtime while guaranteeing \ed-optimality is to take runtime samples with timeout $\frac{\theta}{\delta}$ and reject any algorithm that times out for any instance. Then, a concentration bound on the measurements could be used to ensure that the mean is close to the empirical mean. If the mean is less than $\theta$, Markov's inequality can be used to bound the tail probability for \ed-optimality. However, by rejecting any configuration that ever times out, we fail to measure the capped mean--which could be significantly lower--, and thus the algorithm may not stop at the right time. To fix this, we would ideally allow a $\delta$ fraction of runs to time out, but we use $\frac{3}{4}\delta$ instead, to achieve a high-confidence tail bound with a Chernoff bound (replacing Markov's inequality). Still, the measurements could take a long time: if we perform $b$ measurements for a reliable mean estimation with timeout $\tau$, then we spend up to $b\tau$ time. A key observation is that if we spend more than $b\theta$ time on measurements, the average would have to be above $\theta$, and we would reject the configuration. Thus, we can specify an overall time budget of $T=b\theta$, and reject any configuration early if they run over this limit. These ideas are embodied in our algorithm (\RuntimeEst).

\begin{algorithm}[tbh]
\caption{The \RuntimeEst subroutine}\label{slave-alg}
\begin{algorithmic}[1]
\Inputs{
  Configuration $i$\\
  Instance list $\Jlist=(J_1,\dots,J_b)$ of length $b$\\
  Quantile parameter $\delta\in(0, 1)$\\
  Average runtime bound $\theta$\\
}
\Initialize{
  $T \gets b\theta$ \Comment{overall runtime budget}\\
  $\tau \gets \frac{4\theta}{3\delta}$ \Comment{individual runtime budget}\\
  $j\gets 1$ \Comment{instance index}
}
\While{True} 
  \State Run configuration $i$ on $J_j$ with timeout $\min\{T, \tau\}$
  \State $Q_j\gets R(i, J_j, \min\{T, \tau\})$ \label{alg:rte:qj}
  \State $T\gets T-Q_j$
  \State // Stopping rules:
  \Indent
    \If{$T=0$} \Comment{Stop if overall budget zero} \label{alg:rte:stop}
      \State \Return $\theta$
    \ElsIf{$j=b$} \Comment{Stop after $b=|\Jlist|$ samples} 
      \State \Return $\bar{Q}=1/b\sum_{m=1}^b Q_m$ \Comment{Return mean}
    \EndIf \label{alg:rte:stope}
  \EndIndent
  \State $j\gets j+1$
\EndWhile
\end{algorithmic}
\end{algorithm}

\section{Theoretical Analysis}
\label{sec:analysis}

In this section we explore the theoretical properties of \LeapsAndBounds.
We show that the estimates computed by the algorithm are reliable with high probability. Then we prove that if the estimates are reliable, the running time cannot be too large and the algorithm returns an \ed-optimal solution. 
We start with a few important observations about \RuntimeEst.

\subsection{Guarantees for algorithm \RuntimeEst}

Consider the execution of \RuntimeEst with the inputs $(i,\Jlist, \theta,b)$.
Noting that the loop is stopped if the budget $T_0=b\theta$ gets exhausted, it follows that the total runtime of the (optimized) algorithm is bounded by $b\theta$:
\begin{lemma}\label{slave-runtime}
The runtime of one call to \RuntimeEst 
is $\ordo(b \theta)$.
\end{lemma}

With $T_0=b \theta$, for $j\ge 1$, define $T_{j} = T_{j-1} - Q_j = b\theta - (Q_1+\dots+Q_j)$.
If the budget $b \theta$ is not exhausted (i.e., $T_b=b\theta - (Q_1+\dots+Q_b) >0$),
each instance $J_j$ runs within its $\min\{T_{j-1},\tau\}\le \tau$ individual budget, 
and so $Q_{j}=R(i,J_j)=R(i,J_j,\tau)=:R_j$. Clearly, $T_b>0$ is equivalent to $\bar{Q}<\theta$.
Furthermore, in any case, $Q_j =R(i,J_j,\min(T_{j-1},\tau))\le R_j$.
Defining $\bar R=(R_1+\dots+R_b)/b$, we can summarize these findings as follows:

\begin{lemma}\label{no-tau-used}
If\, \RuntimeEst returns with $\bar{Q}{<} \theta$, then $\forall j, Q_{j}{=}R_{j}$ and $\bar{Q}{=}\bar{R}$. 
Otherwise, $\forall j, Q_{j}{\leq} R_{j}$ and $\bar{Q}{\leq}\bar{R}$.
\end{lemma}
Let us now turn to analyzing \cref{master-alg}.
For this, we need some extra notation.

\subsection{Notation}

Let $\theta_k$, $\tau_k$ and $b_k$ denote the respective values of $\theta$, $\tau$ and $b$ in phase $k$ (\cref{alg:master:sample} of \cref{master-alg}), noting that $\tau_k=\frac{4\theta_k}{3\delta}$.
Let $J_j$ denote the $j$th instance (ever) sampled in \cref{alg:master:sample} of \cref{master-alg}.
Note that for $k$ large enough so that $b_k\ge j$, $J_j$ is the $j$th instance that is passed on to \RuntimeEst by \cref{master-alg} in phase $k$ (for any configuration $i$).
Let $R_{i,j,k}=R(i, J_j, \tau_k)$ be the $\tau_k$-capped runtime of configuration $i$ on instance $J_j$ and let $\bar{R}_{i,k}$ be the average of  these values:  $\bar{R}_{i,k}=\frac{1}{b_k} \sum_{j=1}^{b_k} R_{i,j,k}$. Similarly, let $\bar{Q}_{i,k}$ be the return value of algorithm \RuntimeEst in phase $k$ for configuration $i$, which is also the mean of $(Q_{i,j,k})_j$, the runtimes observed at \cref{alg:rte:qj} of \RuntimeEst. Let $\hat{\sigma}_{i,k}^2$ be the empirical variance of $(R_{i,j,k})_j$: $\hat{\sigma}_{i,k}^2=\frac{1}{b_k}\sum_{j=1}^{b_k} (R_{i,j,k}-\bar{R}_{i,k})^2$.

\subsection{Good events}

Let $p_{i,k}=\Pr_{J\sim\Gamma}\left(R(i, J)> \tau_k \right)$ denote the probability that configuration $i$ does not finish on instance $j$ in time $\tau_k$. 
Next we define two events that ensure that the algorithm works well. First, let
\[
E_{1,i,k} = \{\bar{Q}_{i,k}=\theta_k\} \cup \{p_{i,k} \leq \delta \};
\]
if $E_{1,i,k}$ holds then if \cref{master-alg} returns, the probability that the corresponding configuration fails to solve a random task within $\tau_k$ time is small 
(note that $\bar{Q}_{i,k} \le \theta_k$).

The next event guarantees that the average capped running time is close to its expectation: let
\[
E_{2,i,k} = \{|\bar{R}_{i,k}-R_{\tau_k}(i)| \leq C_{i,k}\}
\vspace{-5pt}
\]
with 
\vspace{-1pt}
\[
C_{i,k} = \hat{\sigma}_{i,k} \sqrt{\frac{2\log (\frac{6nk(k+1)}{\zeta})}{b_k}} + \frac{3\tau_k \log (\frac{6nk(k+1)}{\zeta})}{b_k}~.
\vspace{-6pt}
\]

The main result of this section is to show that $E_{1,i,k}$ and $E_{2,i,k}$ hold with high probability for all $i$ and $k$ simultaneously:
\begin{lemma}\label{no-errors}
Let $E=\bigcap_{i \in \{1, \ldots, n\}, k \in \mathbb{Z}_+} \left(E_{1,i,k} \cap E_{2,i,k}\right)$.
Then, $\Pr(E) \geq 1-\zeta$.
\end{lemma}

To prove the lemma, we individually bound the probabilities that the events do not hold:
\begin{lemma}\label{type1-error-bound}
$\Pr(E_{1,i,k}^c)\leq \frac{\zeta}{2nk(k+1)}$.
\end{lemma}
\vspace{-0.4cm}\begin{proof}
If $p_{i,k}\leq \delta$, then $\Pr(E_{1,i,k}^c)=0$ and the statement holds trivially. For the rest of this proof, we assume that $p_{i,k}>\delta$.
From the algorithm, we have that $b_k\geq \frac{32}{\delta}\log(\frac{2nk(k+1)}{\zeta})$.  
Define $B_{i,j,k}$ as the Bernoulli random variable indicating whether configuration $i$ on input $J_j$ takes
more time than $\tau_k$ to finish (value 1), or not (value 0). For $\hat{\delta}_{i,k}=\frac{1}{b_k}\sum_{j=1}^{b_k}B_{i,j,k}$, observe that $\E(\hat{\delta}_{i,k}) = p_{i,k}$.
If the algorithm returns with $\bar{Q}_{i,k}< \theta$, as necessary for event $E_{1,i,k}^c$, then $\bar{R}_{i,k}=\bar{Q}_{i,k}$ according to \cref{no-tau-used}. Noting that $B_{i,j,k}=\mathbb{I}\left[ R_{i,j}\geq \tau_k\right]$, we have $\frac{4\theta}{3\delta} \sum_j B_{i,j,k} \leq \sum_j R_{i,j}$ (since $\tau_k=\frac{4\theta}{3\delta}$). Therefore,
$\frac{4\theta}{3\delta} \hat{\delta}_{i,k}\leq \bar{R}_{i,k}=\bar{Q}_{i,k}<\theta$, so $\hat{\delta}_{i,k}\leq \frac{3}{4}\delta$.

Applying a Chernoff bound on the $b_k$ independent Bernoulli random variables $B_{i,j,k}$, the probability of the latter event can be bounded, giving
\begin{align*}
\Pr(E_{1,i,k}^c) &= \Pr(\bar{Q}_{i,k}<\theta) \leq \Pr\left(\hat{\delta}_{i,k} \leq \frac{3}{4}\delta\right)\\
&\leq \Pr\left(\hat{\delta}_{i,k}\leq\frac{3}{4}\E(\hat{\delta}_{i,k})\right)
\leq \exp\left(- \frac{\E(\hat{\delta}_{i,k})b_k}{32} \right)\\
&< \exp\left(-\frac{1}{32}\delta  b_k\right)\leq \frac{\zeta}{2nk(k+1)},
\end{align*}
where the second and second to last inequalities follow from $\E(\hat{\delta}_{i,k}) > \delta$. 
\end{proof}

\begin{lemma}\label{type2-error-bound}
$\Pr(E_{2,i,k}^c) \leq \frac{\zeta}{2nk(k+1)}$.
\end{lemma}
\vspace{-0.4cm}\begin{proof}
The samples $\left(R_{i,j,k}\right)_j$ are independent and identically distributed with mean $\bar{R}_{i,k}$ and expectation $R_{\tau_k}(i)$. Thus, the lemma holds by the empirical Bernstein bound 
(cf. \citealp[Theorem~1]{audibert2009} and
\cref{bernstein-appendix}). 
\end{proof}

Now \cref{no-errors} follows from \cref{type1-error-bound,type2-error-bound} and the union bound (details are given in \cref{no-error-appendix}).

\subsection{Bounding the average runtime}
Note that when the algorithm finishes, $\bar{Q}_{i,k}=\bar{R}_{i,k}$.
Hence, in this section we focus on $\bar{R}_{i,k}$ and its deviation from its mean.
In particular, we show that 
$|R_{\tau_k}(i) - \bar{R}_{i,k}| \le \frac{3}{7} \epsilon R_{\tau_k}(i)$ 
holds on event $E$ when phase $k$ is \emph{preterm}. 
Here, a phase $k$ is called preterm if $\min_i R_{\tau_k}(i)\geq \frac{7}{16}\theta_k$. 
The idea is that if a phase is preterm then the best capped expected runtime is large compared to the guess 
on the optimal runtime. We then show that on $E$, any phase executed by the algorithm is preterm.

Since on $E$, $|R_{\tau_k}(i) - \bar{R}_{i,k}|\le C_{i,k}$ by the definition of $E$, 
we need to bound $C_{i,k}$.
We start with a bound on the empirical variance $\hat{\sigma}^2_{i,k}$.
\begin{lemma}\label{var-bound2}
For any preterm phase $k$, 
$\hat{\sigma}_{i,k}^2\leq \frac{32}{21\delta}(\bar{R}_{i,k}+R_{\tau_k}(i))^2$.
\end{lemma}
\vspace{-0.4cm}\begin{proof}
First we show that for any $c>0$, 
\begin{equation}
\label{eq:sig1}
\hat{\sigma}_{i,k}^2\leq c\left(\bar{R}_{i,k}+\frac{\tau_k}{2c}\right)^2~.
\end{equation}
Notice that $\hat{\sigma}_{i,k}^2=\frac{1}{b_k}\sum_{j=1}^{b_k} (R_{i,j,k}-\bar{R}_{i,k})^2 \leq \frac{1}{b_k}\sum_{j=1}^{b_k} R_{i,j,k}^2\leq \frac{1}{b_k}\sum_{j=1}^{b_k}\tau_k\, R_{i,j,k}=\tau_k \, \bar{R}_{i,k}$ because  $\bar{R}_{i,k}  \le \tau_k$ by definition. Now \eqref{eq:sig1} follows from the obvious $\bar{R}_{i,k} \tau_k\leq c\left(\bar{R}_{i,k}+\frac{\tau_k}{2c}\right)^2$.

 By the assumption on $k$, $\theta_k\leq \frac{16}{7}R_{\tau_k}(i)$. Since $\tau_k=\frac{4\theta_k}{3\delta}$, this means that $\tau_k\leq \frac{64}{21\delta}R_{\tau_k}(i)$.  Thus, applying \eqref{eq:sig1} with $c=\frac{32}{21\delta}$ completes the proof as $\hat{\sigma}_{i,k}^2 \leq \frac{32}{21\delta} \left(\bar{R}_{i,k}+\frac{21\delta}{64}\tau_k\right)^2\leq \frac{32}{21\delta}\left(\bar{R}_{i,k}+R_{\tau_k}(i)\right)^2$. 
\end{proof}

Combining the above result with the upper bound $\tau_k=\frac{4\theta_k}{3\delta}\leq \frac{64}{21\delta}R_{\tau_k}(i)$, which holds for any preterm phase $k$, simple algebra yields the following bound on $C_{i,k}$ (the full proof is given in \cref{e-is-good-appendix}):%
\footnote{The multiplicative constant in the proof is not optimized carefully to promote simplicity. Nevertheless, in our experiments the empirical effect of this constant is negligible.}

\begin{lemma}\label{e-is-good}
For any preterm phase $k$, it holds that $C_{i,k} \leq \frac{\epsilon}{3}( \bar{R}_{i,k}+R_{\tau_k}(i))$.
\end{lemma}

Now we give the promised bound on $|R_{\tau_k}(i) - \bar{R}_{i,k}|$.
\begin{lemma}\label{avg-bound}
Assume $E$ holds and $C_{i,k} \leq \frac{\epsilon}{3}( \bar{R}_{i,k}+R_{\tau_k}(i))$. 
Then, $|R_{\tau_k}(i) - \bar{R}_{i,k}|  \le \frac{3}{7}\epsilon R_{\tau_k}(i)$ for all configurations $i$.
\end{lemma}
\vspace{-0.4cm}\begin{proof}
Let us define $x$ such that $\bar{R}_{i,k} = (1+x)R_{\tau_k}(i)$. Because $E_{2,i,k}$ holds, $|x|R_{\tau_k}(i)= |R_{\tau_k}(i)-\bar{R}_{i,k}| \leq C_{i,k} \leq \frac{\epsilon}{3} (\bar{R}_{i,k}+ R_{\tau_k}(i)) =\frac{\epsilon}{3} (1+2x) R_{\tau_k}(i)$. So $|x|\leq \frac{\epsilon}{3} (1+2x)$. If $x < 0$, then $x\geq -\frac{\epsilon/3}{1+2\epsilon/3}>-\frac{3}{7}\epsilon$. If $x\geq 0$, then $x\leq \frac{\epsilon/3}{1-2\epsilon/3}\leq \frac{3}{7}\epsilon$ because $\epsilon\leq \frac{1}{3}$.
\end{proof}

In the analysis of the correctness and the running time of the algorithm, we only need the slightly weaker corollary of \cref{e-is-good} and \cref{avg-bound} (which also holds for another variant of our algorithm, as opposed to \cref{e-is-good}):
\begin{corollary}\label{compatible-bound}
Assume $E$ holds and phase $k$ is preterm. Then, for each $i$, if $\bar{R}_{i,k}<\theta_k$, then $|R_{\tau_k}(i) - \bar{R}_{i,k}|  \le \frac{3}{7}\epsilon R_{\tau_k}\!(i)$; otherwise, if $\bar{R}_{i,k}\!\geq\!\theta_k$, then $\theta_k\!\!<(1\!+\!\frac{3}{7}\epsilon)R_{\tau_k}\!(i)$.
\end{corollary}

\subsection{Correctness and runtime}

In this section we show that Algortihm~\ref{master-alg} returns an \ed-optimal configuration, and give an upper bound on its running time. First we show the following result promised earlier: 

\begin{lemma}\label{induction}
If $E$ holds then every phase $k$ executed is preterm.
\end{lemma}
\vspace{-0.4cm}\begin{proof}
The first phase is preterm as $\frac{7}{16}\theta_1=\kappa_0\leq R_{\tau_1}(i)$.
For a phase $k\geq 2$ that is executed, since the algorithm did not return in phase $k-1$, by \cref{no-tau-used}, $\bar{R}_{i,k-1}\geq \bar{Q}_{i,k-1}=\theta_{k-1}$. 
If $E$ holds and phase $k-1$ was preterm, by \cref{compatible-bound}, $\theta_{k-1}<(1+\frac{3}{7}\epsilon)R_{\tau_{k-1}}(i)$. 
Moreover,
$$\textstyle \frac{7}{16}\theta_{k}= \frac{7}{8}\theta_{k-1}\leq \frac{7}{8} (1+\frac{3}{7}\epsilon) R_{\tau_{k-1}}\!(i)\leq R_{\tau_{k-1}}\!(i)\leq R_{\tau_{k}}\!(i),$$
since $\epsilon\leq \frac{1}{3}$. By induction, any phase executed is preterm.
\end{proof}

\begin{lemma}\label{correctness}
If $E$ holds and \cref{master-alg} returns with a configuration $I$ in phase $K$, then $I$ is \ed-optimal.
\end{lemma}
\vspace{-0.4cm}\begin{proof} 
We prove the statement by contradiction.
Thus, assume $I$ is \ed-suboptimal.
At stopping, $\bar{Q}_{I,K}<\theta_K$, hence on $E$, $p_{I,K} =\Pr_{J\sim\Gamma}(R(I, J)>\tau_K)\leq \delta$ must hold. Since $I$ is \ed-suboptimal, it follows that there exists an instance $j$ such that $R_{\tau_K}(I) > (1+\epsilon) R(j) > (1+\epsilon)R_{\tau_K}(j)$. Take such an index $j$.
Since \cref{master-alg} returned $I$ instead of $j$, $\bar{Q}_{I,K}\le \bar{Q}_{j,K}$. 
By \cref{no-tau-used}, $\theta_k>\bar{Q}_{I,K}=\bar{R}_{I,K}$ and $\bar{R}_{j,K}\geq \bar{Q}_{j,K}$. Applying \cref{compatible-bound} and \cref{induction}, if $\bar{R}_{j,K}<\theta_k$, then $(1+\frac{3}{7}\epsilon)R_{\tau_K}(j)\geq \bar{R}_{j,K}\geq \bar{Q}_{j,K}\geq \bar{Q}_{I,K}$. Otherwise, $(1+\frac{3}{7}\epsilon)R_{\tau_k}(i)\geq\theta_k\geq\bar{Q}_{I,K}$. Using this,
\begin{align*}
R_{\tau_K}(j)(1+\tfrac{3}{7}\epsilon)&\geq \bar{Q}_{I,K}=\bar{R}_{I,K} > R_{\tau_K}(I) (1-\tfrac{3}{7}\epsilon) \\
&> R_{\tau_K}(j)(1+\epsilon)(1-\tfrac{3}{7}\epsilon).
\end{align*}
Therefore, $1+\frac{3}{7}\epsilon>(1+\epsilon)(1-\frac{3}{7}\epsilon)$ which leads to a contradiction since $\epsilon\leq \frac{1}{3}$.
\end{proof}

\begin{theorem}
\label{thm}
\cref{master-alg} identifies an \ed-optimal solution in time
$\ordo\left(\opt \frac{n}{\epsilon^2\delta} \log(\frac{n\log \opt}{\zeta})\right)$ with probability at least $1-\zeta$, where $\opt=\min_i R(i)$.
\end{theorem}
\vspace{-0.4cm}\begin{proof}
By \cref{no-errors}, $E$ holds with probability at least $1-\zeta$. The rest of the proof assumes that $E$ holds.

Let $i^*=\argmin_i R(i)$. If $\theta_k\geq (1+\frac{3}{7}\epsilon) \opt\geq (1+\frac{3}{7}\epsilon) R_{\tau_k}(i^*)$, then only the first case of \cref{compatible-bound} can hold. Together with \cref{induction}, we have that $\bar{Q}_{i^*,k}\leq \bar{R}_{i^*,k}\leq(1+\frac{3}{7}\epsilon) R_{\tau_k}(i^*)\leq\theta_k$, so \cref{master-alg} terminates for $\theta_k\geq (1+\frac{3}{7}\epsilon) \opt$. Let the total number of phases of the outer loop of \cref{master-alg} be $L$. Then $L=\ordo(\log \opt)$.

The for loop on \cref{alg:master:for} of \cref{master-alg} adds a factor of $n$ to the runtime. By \cref{slave-runtime}, calling algorithm \RuntimeEst on \cref{alg:master:runtimeest} adds a factor of $b_k \theta_k$ to the runtime. Now $b_k \leq \ceil{\left( 44\log(\frac{6nL(L+1)}{\zeta})\frac{1}{\delta\epsilon^2}\right)}=\ordo\left(\frac{1}{\epsilon^2\delta} \log(\frac{6n\log^2 \opt}{\zeta})\right)=\ordo\left(\frac{1}{\epsilon^2\delta} \log(\frac{6n\log \opt}{\zeta})\right)$, so substituting $\theta_k=\frac{16}{7}\kappa_0 2^k$, the total runtime becomes
\begin{align*}
&\ordo\left(\sum_{k=1}^{\ceil{\log_2 \left((1+\frac{3}{7}\epsilon) \frac{\opt}{\kappa_0}\right)}} \kappa_0 2^k\cdot \frac{n}{\epsilon^2\delta} \log\Bigg(\frac{6n\log \opt}{\zeta}\Bigg)\right) \\
&= \ordo\left(\opt \frac{n}{\epsilon^2\delta} \log\Bigg(\frac{n\log \opt}{\zeta}\Bigg)\right) \,.
\end{align*}
By \cref{correctness}, when the algorithm returns, it returns with an \ed-optimal configuration.
\end{proof}

\section{Optimizing \RuntimeEst}\label{empirical-stopping}
\label{sec:empirical}

Our runtime analysis presented in the previous section used a  worst-case upper bound for $\hat{\sigma}_{i,k}$. Some instances may allow faster runtimes if we modify \RuntimeEst to stop earlier in scenarios where the empirical variance is lower than this worst case bound. 
To do this, building on the approach of \citet{mnih2008empirical}, we 
change the stopping rules of algorithm \RuntimeEst and add two more rules as given in \cref{stopping-alg}.
The code shown here should replace Lines~\ref{alg:rte:stop}--\ref{alg:rte:stope} of \RuntimeEst. 

\begin{algorithm}[!t]
\caption{Stopping rules}\label{stopping-alg}
\begin{algorithmic}[1]
  \State $\bar{Q} \gets \frac{1}{j}\sum_{m=1}^j Q_m$
  \State $\hat{\sigma}^2 \gets \frac{1}{j}\sum_{m=1}^j \left(Q_m - \bar{Q}\right)^2$
  \State $d_{j,k}\gets 4nk(k+1)j(j+1)/\zeta$
  \State $c\gets \sqrt{\hat{\sigma}^2 \frac{2\log(3 d_{j,k})}{j} } + \frac{3\tau\log(3d_{j,k})}{j}$
  \State $\lb\gets \bar{Q} - c$
  \If{$T=0$} \Comment{Stop if overall budget zero}
      \State \Return $\theta$ \label{l:T0}
  \EndIf
  \If{$j=b$} \Comment{Stop after $b=|J|$ samples}
    \State \Return $\bar{Q}$ \Comment{Return mean of $Q$} \label{l:bJ}
  \EndIf
  \If{$(1+\frac{3}{7}\epsilon)\lb \geq \theta$ and $\bar{Q}>\theta$} \Comment{LB too large}
  \label{alg:stop:lbtoolarge}
    \State \Return $\theta$ \label{l:LB}
  \EndIf
  \If{$j\geq \ceil{\frac{32}{\delta}\log d_{j,k}}$ and $c \leq \frac{\epsilon}{3} \left(\bar{Q} + \lb\right)$}
  	\label{alg:stop:smallc}
    \State \Return $\bar{Q}$ \Comment{Return mean of $Q$} \label{l:d}
  \EndIf
\end{algorithmic}
\end{algorithm}

We outline a proof sketch  that this algorithm is still correct and has the same runtime bound. 
We define the running averages in iteration $j$ of \RuntimeEst as $\bar{Q}_{i,j,k}$ and $\bar{R}_{i,j,k}$.
As before, we define an event, as a union of other events, that guarantees that the empirical estimates behave well.  We keep the previously defined events $E_{1,i,k}$ and $E_{2,i,k}$; note that $E_{1,i,k}$ corresponds to the estimate $\bar{Q}_{i,b,k}$. 
However, we need a similar event to $E_{1,i,k}$ for all iterations $j$: $E_{i,j,k}'= \{\bar{Q}_{i,k}\geq\theta_k\} \cup \{p_{i,k} \leq \delta \} \cup \{j<\ceil{\frac{32}{\delta}\log d_{j,k}}\}$.

Let $E=\bigcap_{i \in \{1,\ldots,n\}, j,k\in \mathbb{Z}_+} \left(E_{1,i,k} \cap E_{2,i,k} \cap E_{i,j,k}' \right)$. Similarly to the previous section, it is easy to show that $\Pr(E) \ge 1- 3\zeta/2$. If $E$ holds and the algorithm returns with an average runtime less than $\theta_k$, then $E_{1,i,k}$ and $E_{i,j,k}'$ guarantee that $\Pr(R(i,j)>\tau_k)\leq\delta$ (independently of which stopping condition was activated). Since the original stopping rule is still in place, the runtime of algorithm \RuntimeEst with the additional stopping rules is still $\ordo (b_k \theta_k)$. Similarly, it is easy to verify that \cref{no-tau-used} still holds.

Furthermore, by a slight modification of Theorem~2 of \citet{mnih2008efficient}, one can show that with probability at least $1-\zeta/2$,
$|R_{\tau_k}(i)-\bar{R}_{i,j,k}|\leq c_{i,j,k}$ 
holds for all $i,j,k$, and $c_{i,j,k}\leq \frac{\epsilon}{3} \left(\bar{Q}_{i,j,k} + \lb_{i,j,k}\right)\leq \frac{\epsilon}{3}\left(\bar{R}_{i,j,k} + R_{\tau_k}(i)\right)$ holds\footnote{Here, \cref{no-tau-used} was used additionally in the last inequality.} for all 
\[
\vspace{-0.1cm}
j\geq
C\cdot\max\!\left(\frac{\sigma_{i,k}^2}{\epsilon^2 R^2_{\tau_k}\!(i)},\frac{\tau_k}{\epsilon R_{\tau_k}\!(i)}\right)
\! \left( \log\frac{1}{\zeta'}+ \log\frac{1}{\epsilon R_{\tau_k}\!(i)}\right)\!,
\]
where $C$ is a universal constant, and $\bar{Q}_{i,j,k}\leq\bar{R}_{i,j,k}$. Denote this event by $E'$; then $\Pr(E') \ge 1-\zeta/2$.%
\footnote{The original event behind $E'$ guarantees, via Bernstein's inequality, that the estimates for the means and variances are accurate enough.}

Applying Lemma~\ref{avg-bound}, $E'$ also implies that $$\textstyle |R_{\tau_k}(i)-\bar{R}_{i,j,k}|\leq \frac{3}{7}\epsilon R_{\tau_k}(i).$$
Thus, if $E\cup E'$ holds, then \cref{compatible-bound} holds: 
 if \cref{stopping-alg} returns either because it went through all the $b_k$ samples or because of \cref{alg:stop:lbtoolarge}, then
$|R_{\tau_k}(i)-\bar{R}_{i,k}|\leq \frac{3}{7}\epsilon R_{\tau_k}(i)$, which implies the first part of the corollary; otherwise \cref{stopping-alg} returns in line~\ref{l:bJ}, implying that the algorithm returns with $\theta_k$ and $(1+\frac{3}{7}\epsilon)R_{\tau_k}>\theta_k$. Then the runtime bound and the correctness guarantee of \cref{thm} follows as before.

On the other hand, if the variances of the runtimes over instances are low enough, it is possible to prove an improved runtime bound for the whole algorithm. For $\zeta'=\frac{\zeta}{4nk(k+1)}$, there exists a constant $C$ such that if $j\geq C\cdot \frac{1}{\delta}\left(\log\frac{1}{\delta}+\log\frac{1}{\zeta'}\right)$, then $j\geq \ceil{\frac{32}{\delta}\log d_{j,k}}$ holds. %
Together with the previous lower bound on $j$ and by upper bounding $\tau_k \le \frac{64}{21\delta} R_{\tau_k}(i)$, by the definition of a preterm phase (see \cref{e-is-good}), with probability at least $1-2\zeta$, \RuntimeEst evaluates at most
$$
C\cdot\max\left(\frac{\sigma_{i,k}^2}{\epsilon^2 R^2_{\tau_k}(i)},\frac{1}{\epsilon \delta}, \frac{1}{\delta}\log\frac{1}{\delta}\right)\left( \log\frac{1}{\zeta'}+ \log\frac{1}{\epsilon R_{\tau_k}(i)}\right)
$$
samples in any phase $k$ for configuration $i$ before the stopping conditions on \cref{alg:stop:smallc} are be satisfied. This bound is usually much lower than the previous $b_k=\ceil{44\log\left(\frac{6nk(k+1)}{\zeta}\right)\frac{1}{\delta\epsilon^2}}$: if the variance of runtimes is sufficiently low, this scales as $\epsilon^{-1}$ rather than $\epsilon^{-2}$ (the $\delta^{-1}\log \delta^{-1}$ term is negligible).

\citet{mnih2008empirical} also describe EBGStop, a slightly improved version of empirical Bernstein stopping, which applies Bernstein inequalities to bound the means of an exponentially increasing number of samples. This allows us to effectively replace $\log\frac{1}{\epsilon R_{\tau_k}(i)}$ with $\log\log\frac{1}{\epsilon R_{\tau_k}(i)}$ in the bound presented above. We use this version of the algorithm in our experiments. For completeness, the pseudocode of this version is given in \cref{ebg-pseudo}. 

\setlength{\textfloatsep}{12pt}
\begin{figure}[th]
\centering
	\includegraphics[width=0.37\textwidth]{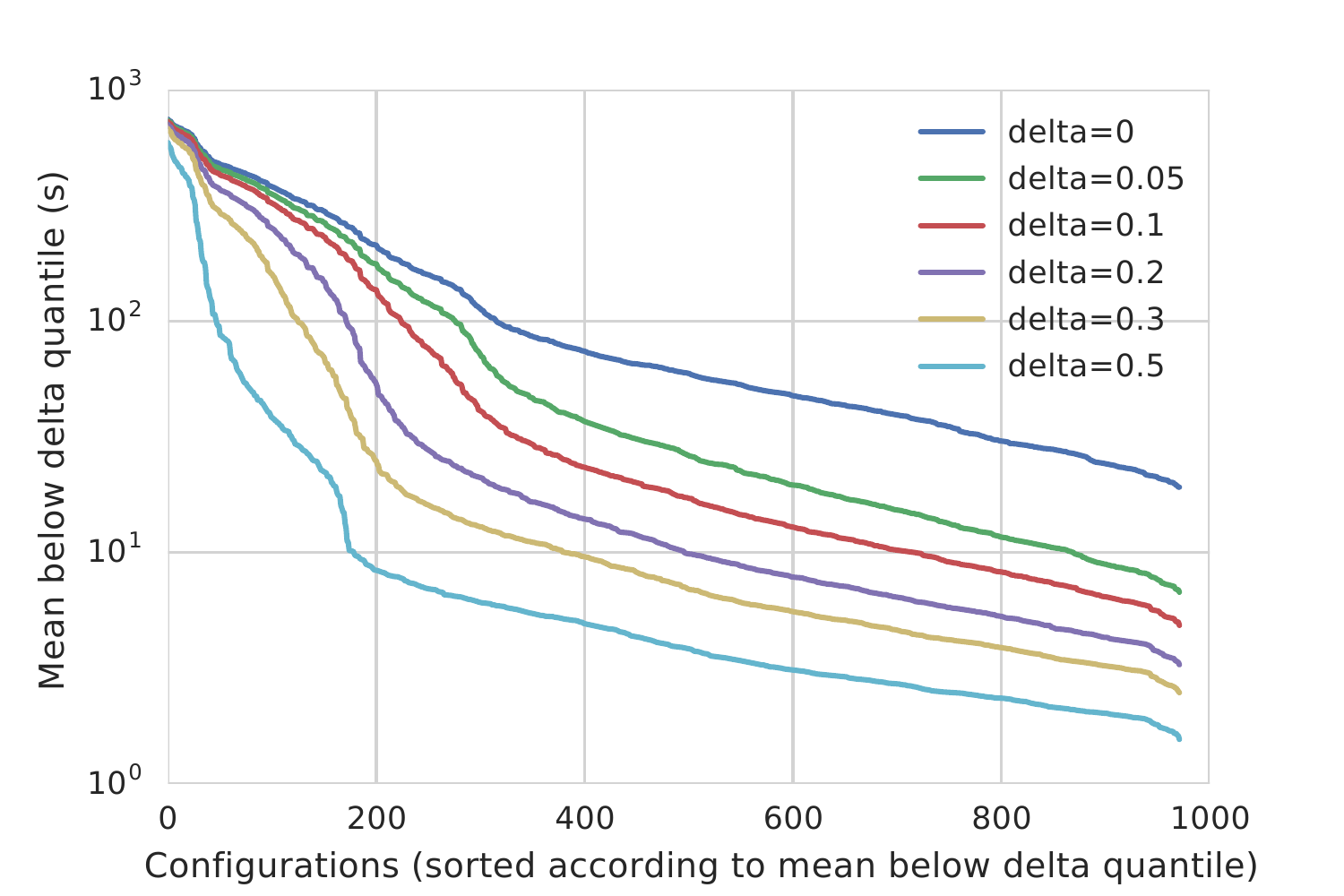}
\caption{Average of runtimes, capped at the timeout, disregarding the worst $\delta$ fraction of samples.
Configurations on the $x$-axis are sorted according to this value.
Note the $\log$ scale on the $y$-axis.
}
\label{mean-below-delta}
\end{figure}
\begin{figure}[tbh]
\vspace{-0.4cm}
\centering
	\includegraphics[width=0.37\textwidth]{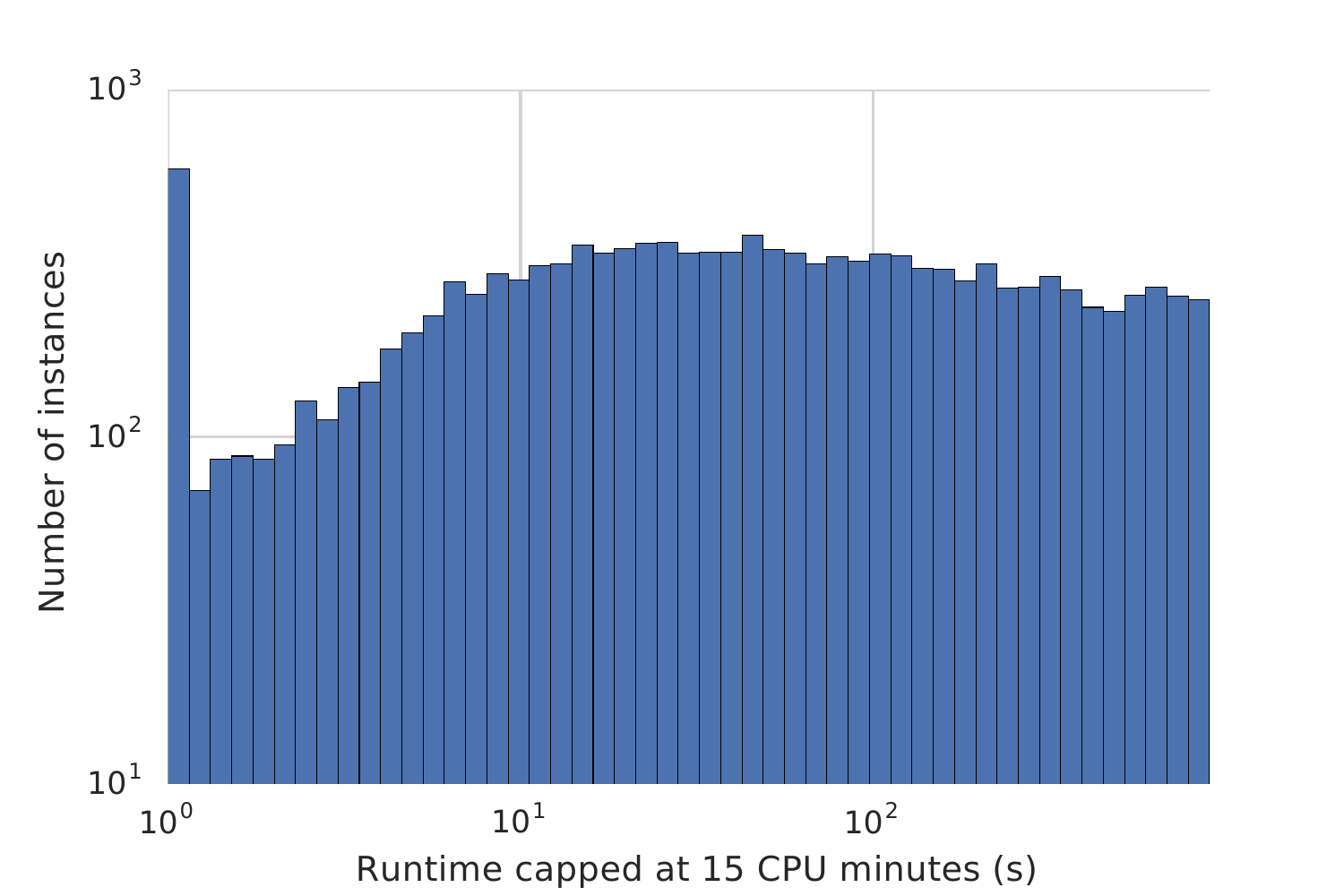}
\caption{Histogram of runtimes of configuration 898 
over the instances in our data. Note the $\log$ scale on the $x$-axis.}
\label{buckets}
\end{figure}

\section{Experiments}
\label{sec:exp}

To run experiments, we gathered a benchmark set of runtimes of different configurations on generated SAT problems. We used \minisat\footnote{We used version 2013/09/25. http://minisat.se/} \citep{sorensson2005minisat} as the SAT solver.
The SAT problems were generated using CNFuzzDD,\footnote{http://fmv.jku.at/cnfuzzdd/} of which only those 20118 were kept that took at least about a second to solve for \minisat with the default parameters. This was done so that the data reflects what happens when instances are nontrivial to solve. 
$972$ different configurations were tested for \minisat, which are described in Appendix~\ref{configurations}. 
The solver \minisat was run with each configuration and instance combination. 
The unit of computation, $\kappa_0$, is one second of CPU time, and the experiments were ran with a timeout of $15$ CPU minutes.\footnote{Our measurements have been scaled such that the unit of computation roughly corresponds to a second on commodity hardware as of $2018$ rather than our machines.
In this unit, about $83$ CPU years were spent in total to generate this data.} 
To get a sense of this data, the capped mean runtimes $R_{\tau}(i)$ 
 for each configuration are shown in
\cref{mean-below-delta} in a sorted order. Here, the timeout $\tau$ was set separately for each configuration so that the tail probability $\Pr_{J\sim\Gamma}(R(i,J)>\tau)$ was approximately $\delta$; the running times are shown for different values of $\delta$ ($\delta=0$ corresponds to the mean runtimes). 
From this, we can see a large difference between configurations. 
For a particularly ``fast'' configuration, \cref{buckets} shows the distribution of runtimes on different instances. Note that because of the global time limit for executions, the final bucket includes runs that may take arbitrarily long.

The benchmark set of runtimes is used to quickly simulate runs of Structured Procrastination and \LeapsAndBounds, as follows. 
A simulated environment acts as an oracle, returning precomputed values of $R(i,j,\tau)$ when queried, accumulating the total time the algorithm under test would have run for. 

Both \LeapsAndBounds and Structured Procrastination often run the same configuration on the same instance with an increased time limit. Thus, both algorithms can benefit largely when the environment allows pausing and resuming of executions. This can be implemented either by saving the state of the execution when the actual runtime limit is reached, or by reloading the state from automatically saved checkpoints. However, resuming execution comes with an additional memory requirement, and may not always be feasible or preferable to restarts. Thus, we report our experiments for both cases.

After each phase, in \cref{alg:master:doubling} of \LeapsAndBounds, we double $\theta$. In fact, this multiplier is arbitrary, and changing it only affects the worst-case runtime up to a constant factor. In practice, a smaller multiplier, making smaller steps in $\theta$, typically overshoots the best average runtime less, thereby decreasing the total runtime for environments that allow resuming runs. On the other hand, a smaller multiplier leads to more phases, introducing more overheads in resuming jobs and increasing the total runtime if resuming is not allowed by rerunning portions of jobs more frequently. The value of the multiplier can be optimized by taking these effects into account, e.g., by measuring the overheads related to switching and resuming jobs. For simplicity, and since this information is not included in our benchmark dataset, in the experiments below the value of the multiplier was set to $1.25$ (see \cref{beta-sweep} for more details). 

\begin{figure}[t]
\centering
	\includegraphics[width=0.37\textwidth]{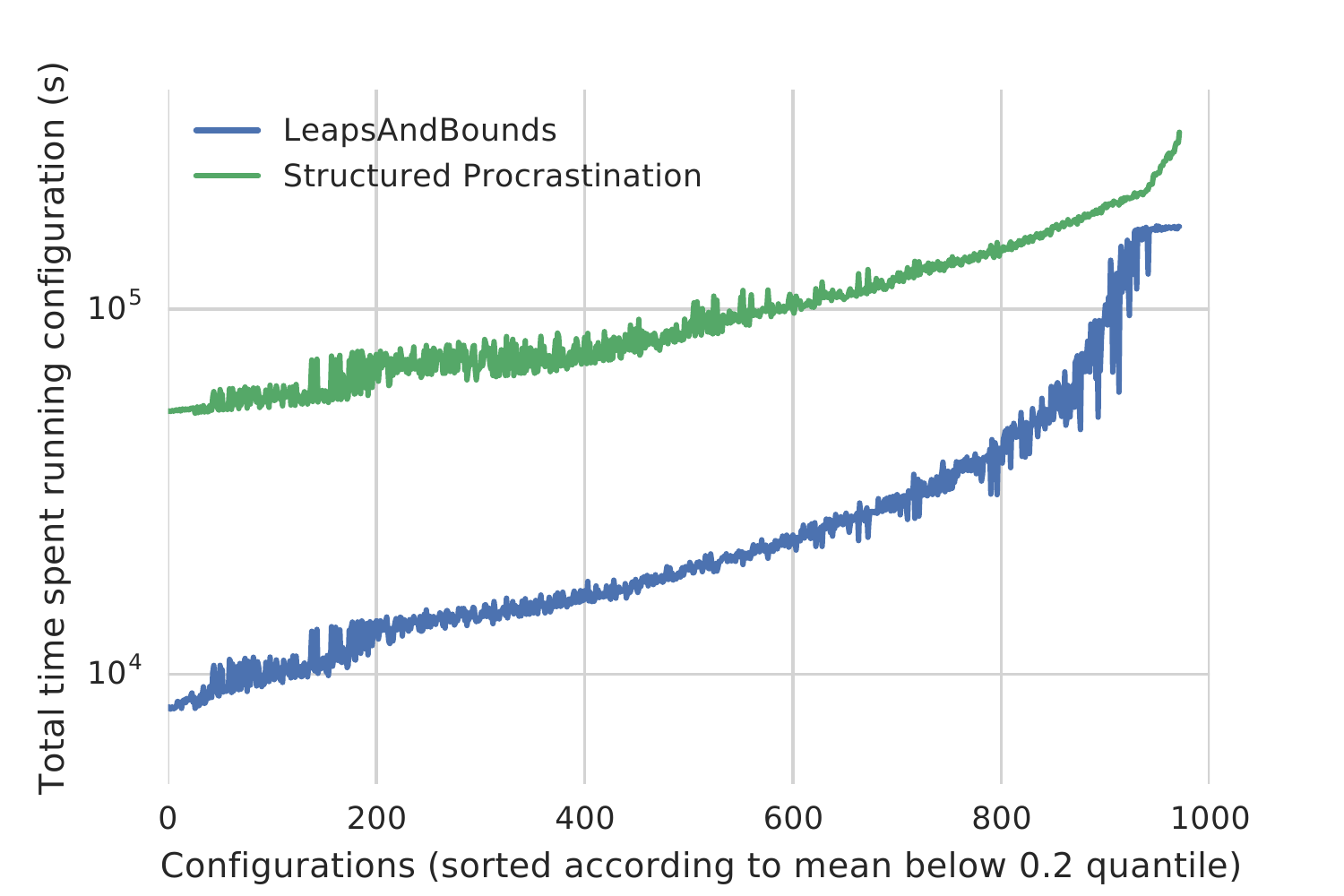}
\caption{Amount of time \LeapsAndBounds runs each configuration compared to Structural Procrastination on a $\log$ scale, in an environment that allows resuming runs.}
\label{time-spent-running}
\end{figure}

We simulated \LeapsAndBounds and Structured Procrastination on our benchmark dataset with parameters $\epsilon=0.2$, $\delta=0.2$, and $\zeta=0.1$.
\cref{time-spent-running} shows that \LeapsAndBounds runs every configuration for a significantly shorter amount of time than Structured Procrastination. The configurations are sorted in the same order as in \cref{mean-below-delta}, for $\delta=0.2$. Paradoxically, both algorithms run the faster configurations significantly longer. This is because both algorithms quickly reject slow configurations, whereas they both run fast configurations many more times to ensure \ed-optimality. In total, \LeapsAndBounds runs for $933.50$ CPU days in the environment that does not support resuming execution, and $368.50$ days in one that does. The corresponding runtime measurements for Structured Procrastination are $1850.46$ and $1169.36$, respectively. Both algorithms return with configuration 898, which has the best average runtime below a $\delta=0.2$ quantile, out of all configurations.

\vspace{-1pt}
\section{Parallelization}
\label{sec:parallel}

One benefit of the simplicity of \LeapsAndBounds is that it is embarrassingly parallel. This is due to the fact that there is little dependency between the runtime measurements that need to be carried out. In phase $k$, when $\theta_k=\frac{16}{7}\kappa_0 2^k$, runs of the form $R\left(i, j, \frac{\kappa_0 2^{k+6}}{21\delta}\right)$ are carried out. The core of our argument is that this parallelizes over $i$, $j$, and $k$, but there are three further considerations. First, to implement the overall runtime bound of \RuntimeEst, for any fixed $i$ and $k$, the runs of $R\left(i, j, \frac{\kappa_0 2^{k+6}}{21\delta}\right)$ should be terminated once the summed running times of these reach the overall budget $b_k\theta_k$. This could be implemented either via inter-process communication or by starting these runs at once on $p$ processors and terminating them after $b_k\theta_k/p$ time.
Second, a new phase $k$ of \cref{master-alg} should only be started once $\bar{Q}_{i,k}$ is available for all $i\in \N$.  Thus, runs should be started in increasing order of $k$, for each $i$. Third, the optional empirical Bernstein stopping, as described in \cref{empirical-stopping}, adds a dependency between runs of different $j$. This could be resolved either by not parallelizing over $j$, or by running only a small number of parallel runs over $j$ and checking the stopping conditions after they finish. 

\vspace{-1pt}
\section{Conclusions and Future Work}
\label{sec:conc}

We have introduced an algorithm applying empirical Bernstein stopping with the goal of finding approximately optimal configurations, and provided guarantees for its worst-case runtime as well as correctness. Our runtime guarantee is tighter than that of Structured Procrastination, which, to our knowledge, is the only other method solving this problem. Empirical evaluations suggest that \LeapsAndBounds outperforms Structured Procrastination in realistic, non-adversarial scenarios too, which depends crucially on leveraging the gap between worst-case and realistic scenarios by using empirical Bernstein stopping. 

The optimality of the configuration returned by \LeapsAndBounds is, in fact, with respect to configurations \emph{with timeout $\tau_K$} for the final phase $K$. An important direction of future work is to get guarantees with respect to the best configuration for the fastest $(1-\delta')$-proportion of instances for any $\delta'<\delta$.

\bibliography{all}
\newpage

\appendix

\section{Empirical Bernstein Bound}\label{bernstein-appendix}

Let $X_1,\ldots X_t$ be i.i.d. random variables with range $R$ and mean $\mu$. Let the empirical mean be $\bar{X}$ and the empirical variance be $\bar{\sigma}^2=\sum_{i=1}^t(X_i-\bar{X})^2$. Applying Bernstein's inequality to the sum and the sum of the squares of these random variables, we get the empirical Bernstein bound \cite{audibert2009}, which states that with probability at least $1-\zeta$,
$$|\bar{X}-\mu|\leq \sqrt{\frac{2\bar{\sigma}^2\log(3/\zeta)}{t}}+\frac{3R\log(3/\zeta)}{t}.$$

\section{Proof of \cref{no-errors}}
\label{no-error-appendix}

From \cref{type1-error-bound,type2-error-bound} and the union bound we obtain
\begin{align*}
\Pr(E) &= 1- \Pr\Big(\bigcup_{i \in \{1,\ldots,n\}, k \in \mathbb{Z}_+} \left(E_{1,i,k}^c \cup E_{2,i,k}^c\right)\Big)\\
&\geq 1-n\sum_{k=1}^\infty \left(\Pr(E_{1,i,k}^c)+\Pr(E_{2,i,k}^c) \right)\\
&\geq 1-\zeta,
\end{align*}
where the last inequality uses \cref{type1-error-bound,type2-error-bound} and the fact that $\sum_{k=1}^\infty\frac{1}{k(k+1)}=1$.  \qed

\section{Proof of \cref{e-is-good}}
\label{e-is-good-appendix}

Using our upper bound on $\hat{\sigma}_{i,k}^2$ from \cref{var-bound2} and that $\tau_k=\frac{4\theta_k}{3\delta}\leq \frac{64}{21\delta}R_{\tau_k}(i)$ for any preterm phase $k$, with the notation $l=\log \left(\frac{6nk(k+1)}{\zeta}\right)$ we have
\begin{align*}
C_{i,k} &= \hat{\sigma}_{i,k} \sqrt{\frac{2l}{b_k}} + \frac{3\tau_k l}{b_k}\\
&\leq
(\bar{R}_{i,k}+R_{\tau_k}(i)) \sqrt{\frac{\frac{64}{21\delta}l}{b_k}} + R_{\tau_k}(i)\frac{\frac{64}{7\delta} l}{b_k}\\
&\leq (\bar{R}_{i,k}+R_{\tau_k}(i)) \left[\sqrt{\frac{\frac{64}{21\delta}l}{b_k}} + \frac{\frac{64}{7\delta} l}{b_k}\right]\\
&\leq (\bar{R}_{i,k}+R_{\tau_k}(i)) \left[\sqrt{\frac{\frac{64}{21\delta}l}{b_k}} + \frac{\frac{64}{21\delta} l}{\epsilon b_k}\right].
\end{align*}
From \cref{master-alg}, we have that $b_k= 44l\frac{1}{\delta\epsilon^2}$. Thus, $\sqrt{\frac{\frac{64}{21\delta}l}{b_k}} + \frac{\frac{64}{21\delta} l}{\epsilon b_k}\leq \frac{\epsilon}{3}$.
\qed

\newpage
\section{Pseudocode of \RuntimeEst with EBGStop Stopping Rules}\label{ebg-pseudo}

\begin{algorithm}[!h]
\caption{The Fast \RuntimeEst subroutine}\label{fast-slave-alg}
\begin{algorithmic}[1]
\Inputs{
  Configuration $i$\\
  Instance list $\Jlist=(J_1,\dots,J_b)$ of length $b$\\
  Quantile parameter $\delta\in(0, 1)$\\
  Average runtime bound $\theta$\\
}
\Initialize{
  $T \gets b\theta$ \Comment{overall runtime budget}\\
  $\tau \gets \frac{4\theta}{3\delta}$ \Comment{individual runtime budget}\\
  $j\gets 1$ \Comment{instance index}\\
  $l\gets 0$ \Comment{logarithmic index}\\
  $\beta\gets 1.1$ \Comment{basis of logarithm}\\
}
\While{True} 
  \State Run configuration $i$ on $J_j$ with timeout $\min\{T, \tau\}$
  \State $Q_j\gets R(i, J_j, \min\{T, \tau\})$ 
  \State $T\gets T-Q_j$
  \State $\bar{Q} \gets \frac{1}{j}\sum_{m=1}^j Q_m$
  \If{$j>\floor{\beta^l}$}
    \State $l\gets l+1$
    \State $\alpha\gets \floor{\beta^l}/\floor{\beta^{l-1}}$
    \State $d'_{l,k}\gets 4\cdot 10.5844 nk(k+1)l^{1.1}/\zeta$
    \State $x\gets \alpha \log (3d'_{l,k})$
  \EndIf
  \If{$T=0$} \Comment{Stop if overall budget zero} 
    \State \Return $\theta$
  \EndIf
  \If{$j=b$} \Comment{Stop after $b=|J|$ samples}
    \State \Return $\bar{Q}$ \Comment{Return mean of $Q$}
  \EndIf
  \If{$j>1$}
   \State $\hat{\sigma}^2 \gets \frac{1}{j}\sum_{m=1}^j \left(Q_m - \bar{Q}\right)^2$
   \State $c\gets \sqrt{2 \hat{\sigma}^2 x/j} + 3 \tau x/j$
   \State $\lb\gets \bar{Q} - c$
   \If{$(1+\frac{3}{7}\epsilon)\lb \geq \theta$ and $\bar{Q}>\theta$} \\ \Comment{LB too large}
     \State \Return $\theta$
   \EndIf
   \State $d_{j,k}\gets 4nk(k+1)j(j+1)/\zeta$
   \If{$j\geq \ceil{\frac{32}{\delta}\log d_{j,k}}$ and $c \leq \frac{\epsilon}{3} \left(\bar{Q} + \lb\right)$}
    \State \Return $\bar{Q}$ \Comment{Return mean of $Q$}
   \EndIf
  \EndIf
  \State $j\gets j+1$
\EndWhile
\end{algorithmic}
\end{algorithm}

\newpage

\section{Configurations for \minisat}
\label{configurations}

The configurations to test for \minisat were the Cartesian product of small sets of values for certain parameters affecting the heuristics of the underlying algorithm. These sets are as follows (note that the default settings are included in the sets): 
\begin{itemize}
    \item{\texttt{'-rinc'}: $[1.1, 2, 5]$,}
    \item{\texttt{'-var-decay'}: $[0.5, 0.95, 0.99]$,}
    \item{\texttt{'-cla-decay'}: $[0.1, 0.5, 0.9, 0.999]$,}
    \item{\texttt{'-rfirst'}: $[10, 100, 1000]$,}
    \item{\texttt{'-phase-saving'}: $[0, 1, 2]$,}
    \item{\texttt{'-ccmin-mode'}: $[0, 1, 2]$.}
\end{itemize}
Configuration 898, which has the best average runtime below a $\delta=0.2$ quantile, corresponds to the following parameter settings: \texttt{'-ccmin-mode=2 -cla-decay=0.999 -phase-saving=0 -rfirst=10 -rinc=5 -var-decay=0.95'}.

\newcommand{\mul}{\gamma}
\section{Choice of the Multiplier of $\theta$}\label{beta-sweep}

\begin{figure}[!h]
\centering
	\includegraphics[width=0.37\textwidth]{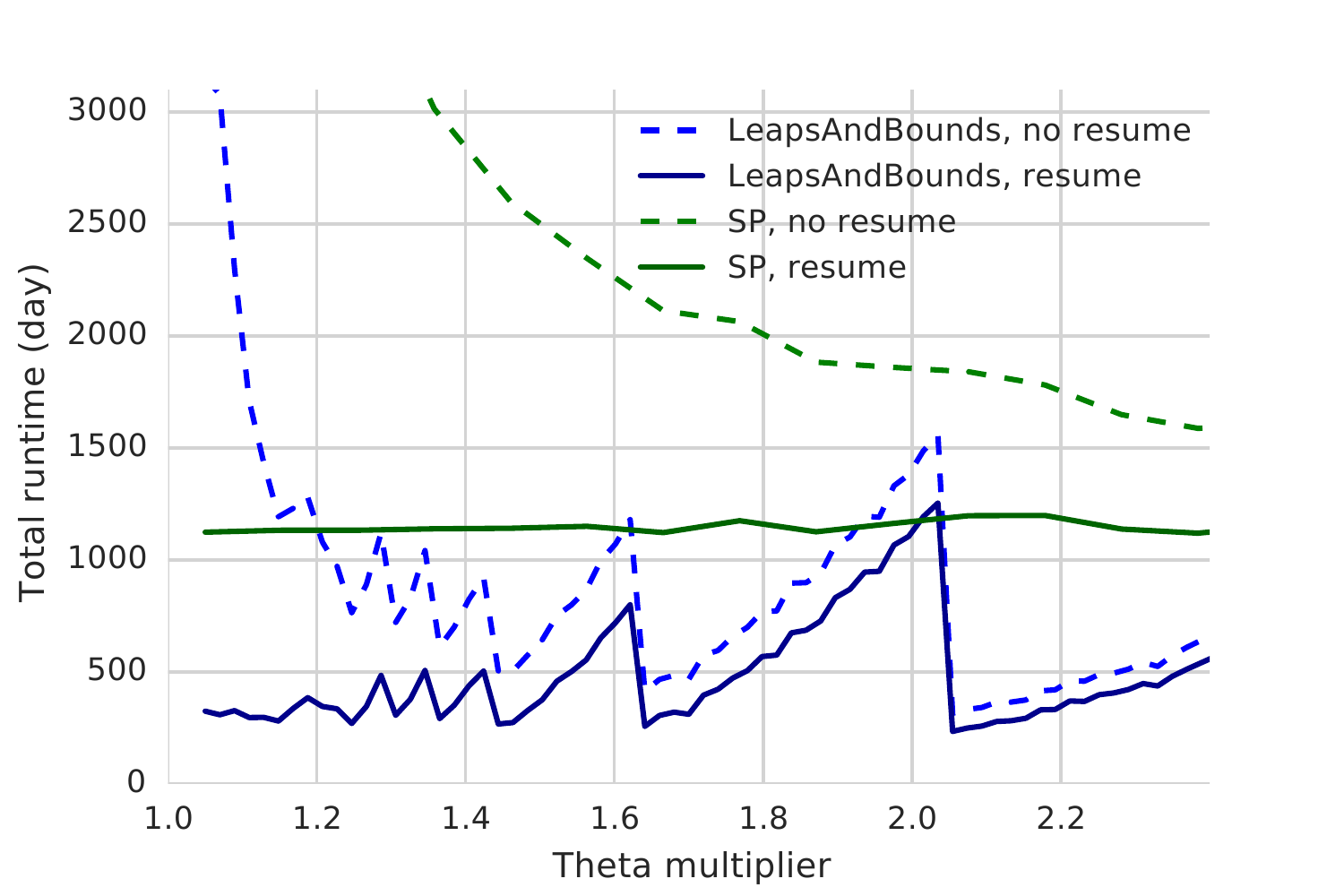}
\caption{Total runtime for different settings of $\theta$ multipliers for \LeapsAndBounds and Structured Procrastination.}
\label{fig:beta-sweep}
\vspace{0.5cm}
\end{figure}

The multiplier of $\theta$, denoted by $\mul$, is also present in Structured Procrastination. \cref{fig:beta-sweep} shows how the total runtime of both methods varies with respect to the multiplier, both in an environment that supports resuming and in one that does not. We can see that Structured Procrastination with resume is not sensitive to the multiplier, while without resume, the total runtime goes up for both methods for a small $\mul$. \LeapsAndBounds outperforms Structured Procrastination for almost any choice of $\mul$ and the environment. The sawtooth-like shape of the runtime curves for \LeapsAndBounds are the result of the overshooting effect: the local minima correspond to cases when the algorithm does not overshoot $\opt$ significantly with $\theta$ in its final phase, and the runtime increases with the amount of the overshoot as $\mul$ increases between two local minima (note that the factor of overshooting is bounded by $\mul$). Our simulations have not modeled the time required to reload a suspended solver for resuming, as this is very environment-dependent. Thus, setting $\mul$ as close to 1 as possible is the best choice when resuming is allowed. In a practical scenario, we suggest that the multiplier $\mul$ be chosen to minimize the upper bound on the runtime, given the cost of reloading.

\balance

\end{document}